\documentclass[letterpaper, 10 pt, conference]{ieeeconf}  % Comment this line out if you need a4paper

\IEEEoverridecommandlockouts                              % This command is only needed if 
\usepackage{amsmath} % assumes amsmath package installed
\usepackage{amssymb}  % assumes amsmath package installed

\usepackage{xspace}

\usepackage{multicol}
\usepackage{multirow}
\usepackage{colortbl}
\usepackage{booktabs} % for rule
\usepackage{bbding} % markers
\newcommand{\ourrow}{\rowcolor{gray!10}}

\newcommand{\ci}[1]{\textcolor{gray}{~($\pm #1$)}}

\def\arrvline{\hfil\kern\arraycolsep\vline\kern-\arraycolsep\hfilneg}
\usepackage{arydshln}

\usepackage{adjustbox}
\usepackage[font=footnotesize,labelfont=normalfont]{caption}

\usepackage[dvipsnames]{xcolor}
\definecolor{citecolor}{HTML}{3953A4} %3953A4, 0071bc
\definecolor{lblue}{HTML}{0071bc} %3953A4
\definecolor{ogreen}{HTML}{2E7D32}
\definecolor{bred}{HTML}{BF360C}
\definecolor{newbrown}{HTML}{795548}

\usepackage{hyperref}
\makeatletter
\let\NAT@parse\undefined
\makeatother
\usepackage[numbers,sort,compress]{natbib}
\usepackage[all=normal,bibbreaks=tight,paragraphs=tight,floats=tight]{savetrees}
\hypersetup{
colorlinks=true, linkcolor=bred, urlcolor=lblue, citecolor=citecolor,
}

\newcommand{\mc}[1]{\mathcal{#1}}

\newcommand{\algoName}{ViSkill\xspace}

\title{\LARGE \bf Value-Informed Skill Chaining for Policy Learning of \\ Long-Horizon Tasks with Surgical Robot}

\author{Tao Huang$^*$, Kai Chen$^*$, Wang Wei, Jianan Li, Yonghao Long, Qi Dou% <-this % stops a space
\thanks{T. Huang, K. Chen, W. Wei, J. Li, Y. Long, and Q. Dou are with the Department of Computer Science and Engineering, The Chinese University of Hong Kong. The * indicates equal contribution.}
\thanks{This work was supported in part by Shenzhen Portion of Shenzhen-Hong Kong Science and Technology Innovation Cooperation Zone under HZQB-KCZYB-20200089, in part by Hong Kong Research Grants Council Project No. 24209223, in part by Science, Technology and Innovation Commission of Shenzhen Municipality Project No. SGDX20220530111201008, and in part by Hong Kong Multi-Scale Medical Robotics Center.}
\thanks{Corresponding author: Qi Dou (qidou@cuhk.edu.hk).}%
}

\begin{document}

\maketitle
\thispagestyle{empty}
\pagestyle{empty}

%%%%%%%%%%%%%%%%%%%%%%%%%%%%%%%%%%%%%%%%%%%%%%%%%%%%%%%%%%%%%%%%%%%%%%%%%%%%%%%%
\begin{abstract} 
Reinforcement learning is still struggling with solving long-horizon surgical robot tasks which involve multiple steps over an extended duration of time due to the policy exploration challenge. Recent methods try to tackle this problem by skill chaining, in which the long-horizon task is decomposed into multiple subtasks for easing the exploration burden and subtask policies are temporally connected to complete the whole long-horizon task.
However, smoothly connecting all subtask policies is difficult for surgical robot scenarios.
Not all states are equally suitable for connecting two adjacent subtasks. 
An undesired terminate state of the previous subtask would make the current subtask policy unstable and result in a failed execution.
In this work, we introduce value-informed skill chaining (ViSkill), a novel reinforcement learning framework for long-horizon surgical robot tasks. 
The core idea is to distinguish which terminal state is suitable for starting all the following subtask policies.
To achieve this target, we introduce a state value function that estimates the expected success probability of the entire task given a state.
Based on this value function, a chaining policy is learned to instruct subtask policies to terminate at the state with the highest value so that all subsequent policies are more likely to be connected for accomplishing the task. 
We demonstrate the effectiveness of our method on three complex surgical robot tasks from SurRoL, a comprehensive surgical simulation platform, achieving high task success rates and execution efficiency. Code is available at \url{https://github.com/med-air/ViSkill}.
\end{abstract}
\section{Introduction}

\begin{figure}[t]
    \centering
    \vspace{0.25cm}
    \centerline{\includegraphics[width=1\linewidth]{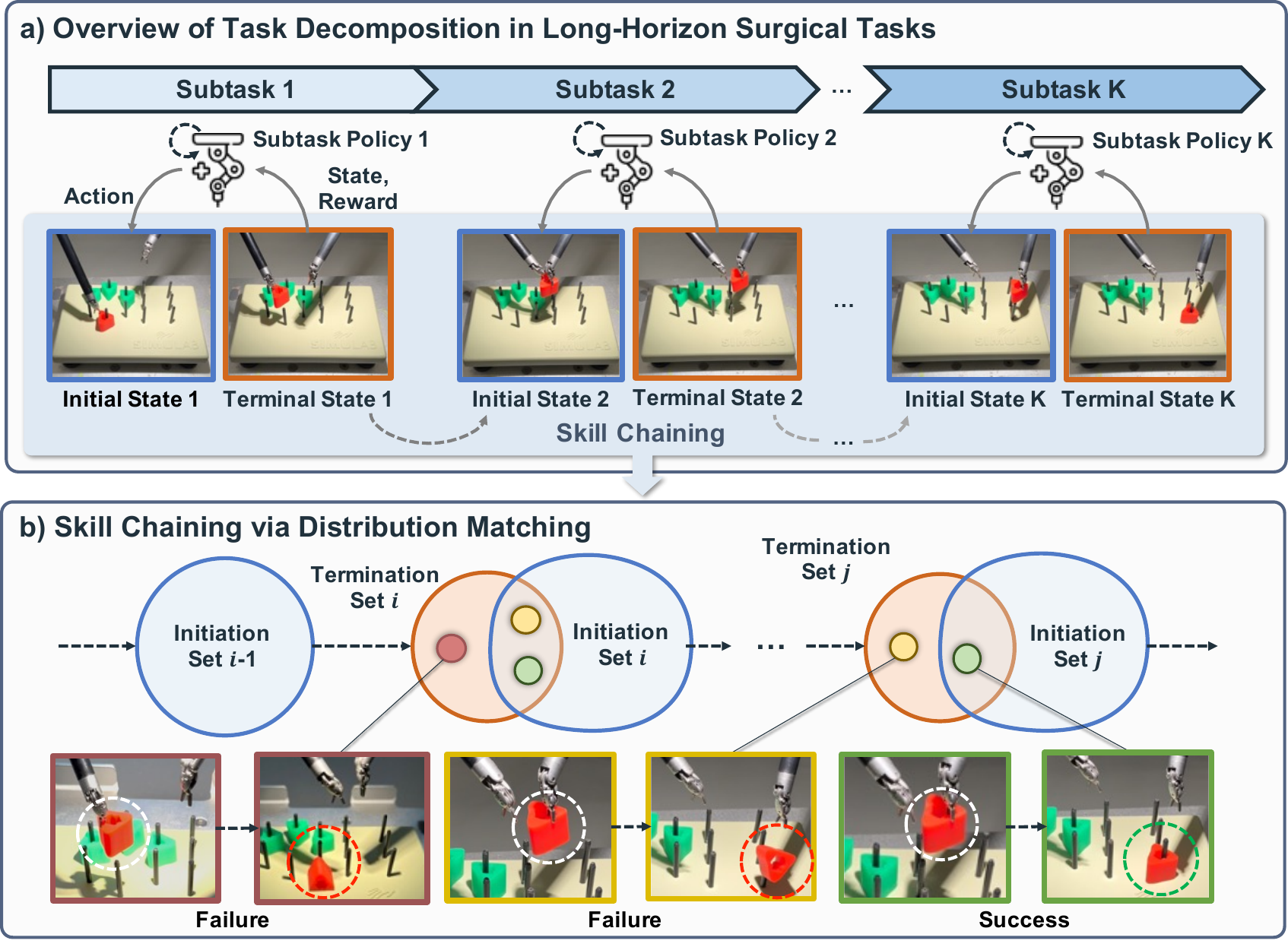}}
        \caption{\textbf{Policy learning for long-horizon surgical tasks.} (a) We decompose the entire long-horizon surgical tasks into multiple subtasks, learn a distinct RL policy (skill) for each subtask, and connect these subtask policies to accomplish the whole task. (b) Prior skill chaining methods take a coarse distribution matching to connect present subtask policies, which may fail to connect subsequent skills due to the sensitivity to the state variation.} 
    \label{fig:overview}
    \vspace{-0.6cm}
\end{figure}

Learning-based surgical robot automation has been increasingly investigated in recent years, with its potential to improve the precision and efficiency of surgical tasks~\cite{ras}.
As is known, robotic surgical tasks are typically long-horizon and composed of several sub-steps with a series of actions over an extended period of time~\cite{meli2021autonomous}.
Such tasks usually have complex specifications, involving sequential operations on small objects. For instance, even in a simplified scenario of basic skill training, automating the bimanual peg transfer task is still challenging, which requires the robot to pick the block up, hand it over to another manipulator, and finally place it to the target peg without collision (see Fig.~\ref{fig:overview}a). 
To date, how to effectively learn the control policy for long-horizon tasks via reinforcement learning (RL) is still an open challenge in the field of surgical robot learning.

Existing RL methods still have limitations when addressing long-horizon tasks, because the agent requires extensive data collection to explore useful behaviors in a large state space, and needs many optimization steps to learn multiple skills with sparse reward signals.
Given these issues, existing methods often resort to considerable reward designing~\cite{ibarz2021train} to facilitate both exploration and policy learning. Unfortunately, these approaches usually require subjective manual engineering, which can result in undesired behaviors~\cite{riedmiller2018learning} and make the policy stuck in local optima~\cite{ddpgher}.
In addition, the cost of manual engineering would become unaffordable when the complexity of the task increases.

A more practical solution is to decompose the long-horizon task into a sequence of easier-to-learn subtasks.
In this way, each subtask is associated with a single action so that the burdens of both exploration and policy learning will be reduced. 
Then, separate control policies are learned to master the distinct skill corresponding to each subtask and all skills are sequentially executed to complete the whole long-horizon task.
Nevertheless, naively executing one policy after another would fail, because the terminal state of one subtask policy is not necessarily able to be handled by the next subtask policy~\cite{lee2020learning}.
Taking the bi-manual peg transfer task in Fig.~\ref{fig:overview} as an example, it can be split into three subtasks \textit{`block picking'}, \textit{`block handing over'}, and \textit{'block placing'}. 
If the hand-over position executed by the left robot arm is not reachable for the right robot arm, it is impossible for the right arm to complete the subsequent subtask of \textit{`block placing'}.
Connecting the subtask policies for accomplishing a long-horizon task is non-trivial, which gives rise to new skill chaining algorithms.

To ensure smooth connections, some prior methods have been proposed to learn transition policies~\cite{lee2019composing,byun2022training}, which aim to transit the agent from terminal states (\textit{i.e.,} termination set) of the subtask to the initial states (\textit{i.e.,} initiation set) of the next subtask, so that the next policy is able to accomplish the subsequent subtask. 
While straightforward, the transition between two subtasks would fail when the terminal state is far from the initiation set of the next subtask policy~\cite{lee2022adversarial}. 
Alternatively, another group of approaches attempts to directly force the termination set of one policy to be covered by the initiation set of the next policy through distribution matching~\cite{clegg2018learning,lee2022adversarial}. 
However, constraining subtask terminal states via distribution matching is too coarse.
As a result, the terminal states will gradually deviate from the initial set of the following policy and lead to the failures of future subtasks.
As illustrated in Fig.~\ref{fig:overview}b, a tiny variation of the block pose after picking up makes the block difficult to be handed over, which is aggregated along the subtask sequence and hampers the completion of placing the object to the peg. In other words, not all states are suitable for connecting two adjacent subtasks, and these methods lack an effective mechanism to evaluate the terminal states of each subtask.

In this work, we present \algoName (Value-informed Skill Chaining), a novel RL-based framework for long-horizon surgical robot tasks. 
Instead of connecting subtask policies via coarse distribution matching, our methods evaluate the terminal states of each subtask with a learned state value function, which estimates the expected success probability of the entire task given a state. 
To terminate subtask policies at states with high values, a chaining policy is introduced to instruct each policy with a subgoal at the initial state. 
Consequently, our method is of high accuracy in chaining all subtask policies and accordingly accomplishing the whole task. 
We demonstrate the effectiveness of our method on three long-horizon surgical robot tasks from the open-source surgical simulation platform SurRoL~\cite{surrol}. 
The experiment results empirically show that our method achieves high task success rates and execution efficiency. 
We also deploy the learned policy to the da Vinci Research Kit~(dVRK) hardware platform, which validates the effectiveness of \algoName on the real robot.
Our contributions are summarized as follows:
\begin{itemize}
    \item We propose a novel value-informed skill chaining algorithm, which considers the expected success probability of the entire task when learning a value function to evaluate the terminal state. The state with the highest value is selected for connecting subtask policies smoothly.
    \item Based on our skill chaining idea, we develop a novel RL framework for long-horizon tasks of surgical robots. We empirically validate the effectiveness of our methods in three representative tasks from SurRoL~\cite{surrol}.
    \item We deploy our method to the dVRK platform and demonstrate the feasibility of executing the policy's predicted motion trajectory on the real robot.
\end{itemize}
\section{Related Work}

\subsubsection{Policy learning for surgical automation} Automating surgical robot tasks with learning-based methods, representatively RL, gets increasing attention in the last decade, owing to its advantages in task generalization. However, most of the research has focused on automating elementary subtasks, such as pattern cutting~\cite{cutting1,nguyen2019new,nguyen2019manipulating} and tissue retraction~\cite{flexml,srl_reward,tissue_lfd}. Their task-specific methods are not easily applicable to the task that are long-horizon
and composed of several elementary actions over an extended period of time. Recently, visible attempts have been made to automate such long-horizon surgical tasks, including bimanual peg transfer~\cite{zhang2022human,huang2023guided}, peg-and-ring~\cite{ginesi2019knowledge,ginesi2020autonomous}, suturing~\cite{schwaner2021autonomous,schwaner2021autonomous-suture,srl_multistage,wilcox2022learning}, and tissue manipulation~\cite{meli2021autonomous}. These works mainly leverage learning from demonstrations approach, typically dynamic movement primitives, to learn the motions of each elementary action and compose them to accomplish the entire task. Nevertheless, they require substantial task-specific expertise in specifying the goal position of each subtask and designing the rule-based connections between subtasks, which makes these methods difficult to be developed at scale. In contrast, our method exhibits higher scalability by flexibly learning both subtask policies and achieving smooth connections between them with a value-informed skill chaining approach.

\subsubsection{Skill chaining for long-horizon tasks} Recently, deep reinforcement learning has presented a scalable framework for learning control policies in robotic tasks. However, solving long-horizon tasks with a single, flat RL policy is still challenging due to the exploration burden on large behavior space and typical sparse-reward setting. 
To this end, some skill chaining methods are proposed to explicitly decompose the task into multiple subtasks, learn an individual policy for each subtask, and sequentially execute subtasks policies to perform the entire task. The connections between skills are achieved by constructing skill trees~\cite{konidaris2009skill, konidaris2012robot, bagaria2020option, bagaria2021robustly}, learning transitional policies~\cite{lee2019composing,byun2022training}, and distribution matching~\cite{clegg2018learning,lee2022adversarial}. However, these methods lack an effective mechanism to evaluate the terminal states of each subtask, while the connections between future subtask policies are sensitive to the variation of the terminal state~\cite{ghoshdivide}. In contrast, we propose a value-informed skill chaining method that learns a value function to estimate the states with higher values if starting from them makes the following subtask policies more likely to be connected, thus achieving a higher success rate for the whole task.

\begin{figure*}[t]
    \centering
    \centerline{\includegraphics[width=1\linewidth]{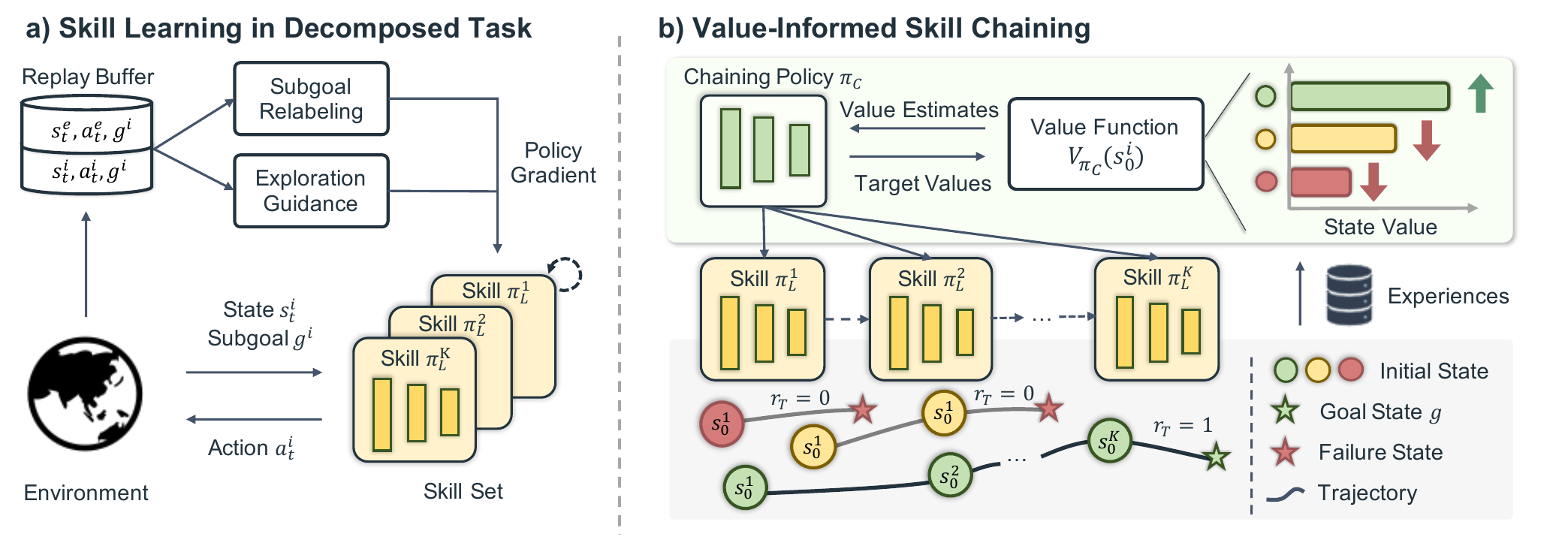}}
    \vspace{-0.2cm}
    \caption{\textbf{Overview of the proposed ViSkill Framework.} In a), we decompose the long-horizon task into multiple subtasks and efficiently learn subgoal-based subtask policies based on subgoal relabeling and demonstration-guided exploration. In b), our value-informed skill chaining method learns a chaining policy that instructs each subtask policy with a subgoal to terminate at states with high state values for smooth connections between skills.}
    \label{fig:method}
    \vspace{-0.4cm}
\end{figure*}

\section{Method}
We introduce ViSkill, a novel reinforcement learning framework for precisely chaining manipulation skills for solving long-horizon surgical robot tasks, as illustrated in Fig.~\ref{fig:method}. Our method first decomposes long-horizon tasks into a sequence of subtasks and learns a subtask policy for each subtask in Section~\ref{method_skill_learning}. To ensure the smooth connections between skills, we subsequently introduce a chaining policy informed with a value function in Section~\ref{method_skill_chaining}.

\subsection{Problem Formulation of Goal-Conditioned RL}\label{method_notations}
We formulate the problem of solving a long-horizon task as a goal-conditioned Markov decision process (MDP) $\mc{M}$~\cite{nasiriany2019planning}. Every episode starts with sampling a goal $g$ from goal space $\mc{G}$ and an initial environment state. The goal stays fixed during the whole episode. At each time step $t$, the agent receives the state $s_t$ and subsequently executes an action $a_t$ to do interactions. The environment transitions to a new state $s_{t+1}$ and yields a $0/1$ sparse reward $r(s_t,a_t,g)$, which is $1$ if the goal is reached at the successor state $s_{t+1}$. An episode terminates after taking $T$ environment steps. The agent aims to learn a control policy that maximizes the expected return $\mathbb{E}[\sum_{t=0}^{T-1}r(s_t,a_t,g)]$.

\subsection{Skill Learning in Decomposed Long-Horizon Tasks}\label{method_skill_learning}
Solving such tasks featuring long-horizon structures and sparse-reward with a single, monolithic control policy is challenging due to its limited capacity to encode and coordinate all required skills. We decompose a long-horizon task into a sequence of subtasks and learn a distinct control policy for each subtask. Specifically, we factorize a single policy into a set of $K$ subtask policies $\pi_L:=\{\pi_L^{1}, \pi_L^{2},...,\pi_L^{K}\}$, where the order of subtasks is indicated by the superscript $i$ and is assumed to be fixed~\cite{andreas2017modular}. Each skill will be instructed with a subgoal $g^i\in\mathcal{G}^i$ at an initial state sampled from its initiation set~$\rho^i$, which is set up by the environment and fixed during the learning process following~\cite{lee2022adversarial}. Each skill agent then interacts with the environment to collect experiences and stores them in its own replay buffer. The objective of each skill is to successfully accomplish the subtask by reaching the instructed subgoal within the subtask episode $T^i$ indicated by a binary reward function $r^i$, \textit{e.g.,} placing the object to a target position.  

 However, learning each skill requires the agent to extensively explore the diverse behaviors in a prohibitively large state space, especially in surgical tasks which may include multiple surgical tools and randomly located small objects. To this regard, we aim to facilitate the exploration by providing each subtask policy with a set of demonstration data. The demonstration-guided RL algorithm DEX~\cite{huang2023guided} is then adopted for skill learning, which encourages the exploration on expert-like trajectories to reduce unproductive behaviors and use subgoal relabelling to enrich the experiences. Specifically, we train each skill by jointly maximizing the expected sum of subtask rewards and behavioral similarity between agent action and expert action $a_t^e$ estimated from subtask demonstration:
\begin{align}\label{eq:subtask_objective}
    \mathbb{E}_\pi\left[\sum_{t=0}^{T^i-1}(r^i(s_t^i,a_t^i,g^i)-\alpha\cdot\mathrm{dist}(a_t^i, a^e_t))\Big|s_0^i\sim \rho^i \right].
\end{align}

While the prior methods depend on an additionally learned reward function~\cite{lee2022adversarial} or a transitional policy~\cite{lee2019composing,byun2022training} to ensure high generalization capabilities to reach different subgoals, the introduction of subgoal-conditioned subtask policies naturally empowers the skill with such ability~\cite{schaul2015universal}. To this end, we train all skills with an initiation set in a limited size and fix them in the later chaining stage, which circumvents the necessity of policy finetuning required by the prior methods given proper subgoal instructions.

\subsection{Skill Chaining via Value-Informed Subtask Termination}\label{method_skill_chaining}
Once acquired skills for each subtask, the naive sequential execution of skills is likely to fail when one skill terminates at the state that the next skill is unable to handle, \textit{e.g.,} the states lie outside the initiation set of the next skill. To overcome this issue, existing methods constrain the termination set of the previous skill to be covered by the initiation set of the next skill. Although such matching between two sets ensures that the next skill will start from its initiation set, the terminal states gradually deviate from the initial set of the following skill due to its sensitivity to the state variation~\cite{ghoshdivide}, leading to the failures of future subtasks. In other words, the connection between current skills does not ensure the successful chaining of all future skills. 

To this regard, we aim to terminate the previous skill in the states, starting from which the agent is not only likely to execute the next subtask successfully, but also is likely to accomplish all remained subtasks. Our key insight is that the chaining of the current two skills should smooth the chaining of all future skills, thus accomplishing the entire task with high probability. Specifically, we first introduce a chaining policy $\pi_C$ to instruct the subtask policies with subgoals. The termination set of one skill, which consists of terminal states of successful subtask executions, is then determined by the chaining policy. To terminate the subtask policy at desired states, we estimate the state value in terms of the whole task reward given an initial state $s^i_0\in\rho^i$ of the subtask, which is defined as follows:
\begin{align}
    V_{\pi_C}(s^i):=\mathbb{E}_{\pi_C,\pi_L}\left[\sum_{j=i}^K R_j|s_0^i=s^i,g^j\sim\pi_C(\cdot|s_0^j)\right],
\end{align}
where $R_i:=\sum_{t=0}^{T^i-1}r(s_t^i,\pi_L^i(s_t^i,g^i),g)$ denotes the subtask return. 
In our sparse-reward settings, the agent only receives a positive reward $r_T=1$ if the final state reaches the goal at the end of the episode (\textit{e.g.}, place the object to the target position successfully). Accordingly, the value function equivalently measures the expected success probability of the whole task given a state:
\begin{align}
    V_{\pi_C}(s^i):=\mathbb{E}_{\pi_C,\pi_L}\left[r_T\right]=\mathrm{Prob}(r_T=1).
\end{align}

At a high level, higher state values of initial states indicate the following subtask policies are more likely to be connected. Terminating the previous subtask policy at such states is thus desired for accomplishing the whole task. To learn such a value function, we minimize the residual Bellman error as follows:
\begin{align}
    \mathbb{E}_{\pi_C,\pi_L}\left[\sum_{i=1}^{K-1} ((R_i+V_{\pi_C}(s_{0}^{i+1})-Q_{\pi_C}(s_0^i,g^i))^2\right],
\end{align}
where $Q_{\pi_C}$ denotes the Q-value function, and the raw environment reward is used as the state value in the last subtask in practice, as the subgoal of the last subtask policy is equal to the task goal. This value function bears resemblances to the one proposed in~\cite{bagaria2021robustly}, while ours measures the success probability of the entire task instead of each subtask. 

Based on the value function, the chaining policy then instructs the subtask policy with a subgoal to terminate at states with high state values for ensuring the smooth chaining of all future skills. It can be learned through the policy gradient method with the algorithm-specific actor loss $\mathcal{L}_{actor}$. While the sparse reward function may incur sample inefficiency given the large manipulation space, we adopt the self-imitation method~\cite{oh2018self} to augment the actor loss for encouraging exploration. In our context, it essentially learns to imitate the successful trajectories $\mathcal{D}_{sil}$ that ends at the state in each termination set, which gives the final objective of the policy optimization:
\begin{align}
\mathcal{L}_{actor} + \lambda\cdot\mathbb{E}_{\mathcal{D}_{sil}}\left[ \sum_{i=1}^{K-1}\log\pi_C(g^i|s^i) w(s^i_0,g^i) \right],
\end{align}
where $\lambda$ is a temperature coefficient, and the weight function $w(s^i,g^i)$ is set as the advantage function following~\cite{awac}. By iteratively updating the value function and chaining policy under off-the-shelf RL algorithms, the subtask policies can be smoothly connected for accomplishing the entire task.

\begin{figure}[t]
    \centering
    \vspace{0.15cm}
    \centerline{\includegraphics[width=1\linewidth]{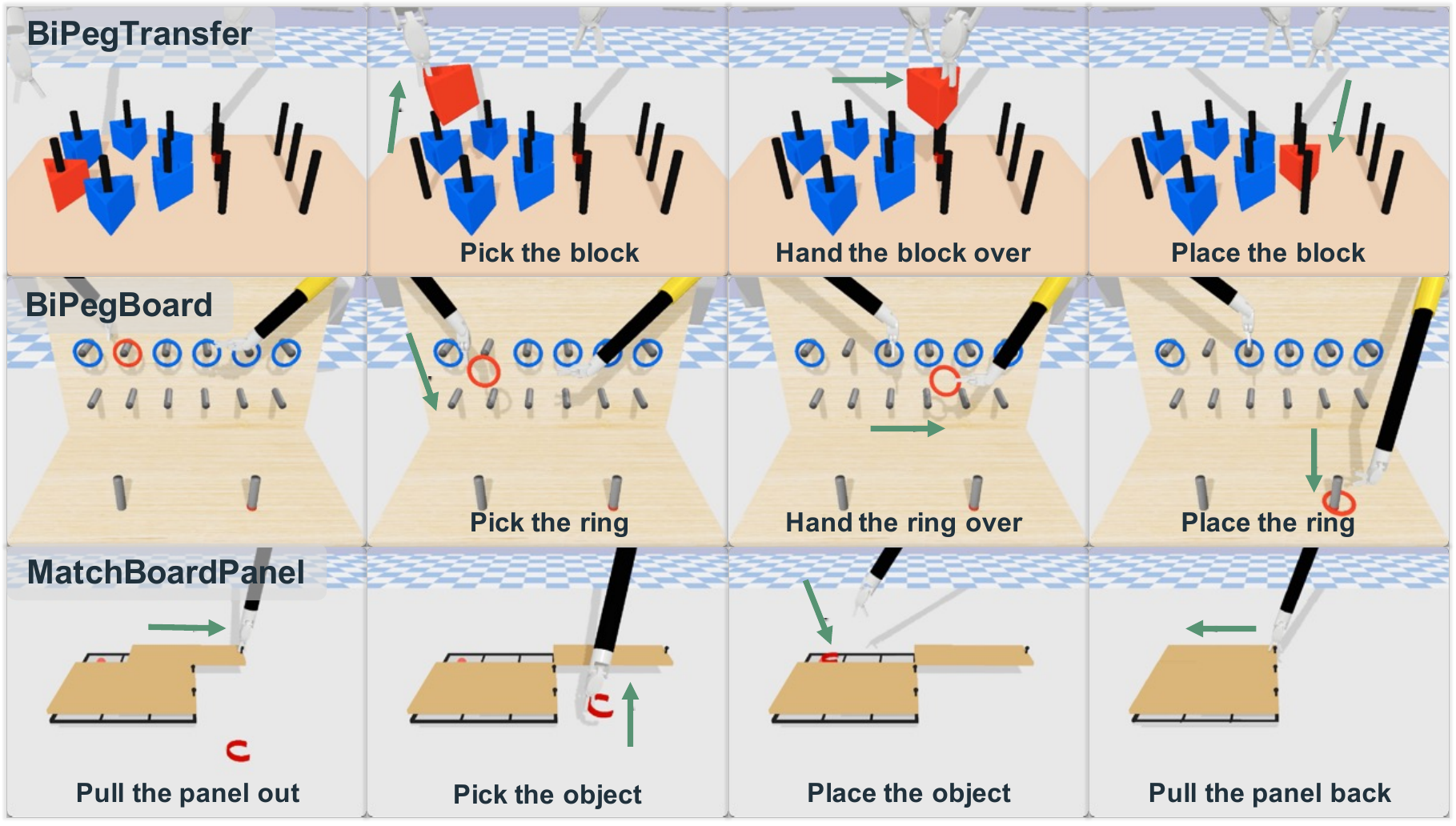}}
    \caption{\textbf{Task descriptions.} Three long-horizon tasks are selected from the surgical robot learning platform SurRoL. The \textit{BiPegTransfer} task and \textit{BiPegBoard} task are composed of three subtasks and the \textit{MatchBoardPanel} is composed of four. Scripted controllers are used to collect subtask demonstrations whose execution flow follows the above snapshots.}
    \label{fig:task}
    \vspace{-0.4cm}
\end{figure}

\begingroup
\setlength{\tabcolsep}{3.5pt}
\begin{table*}[t]
    \centering
    \caption{\textbf{Overall performance comparison.} We present a performance comparison between our method and baselines with respect to the success rate of the entire task, the number of completed subtasks (subtask completion), and the steps taken to complete the task in successful episodes (rollout length). The means and standard variation are reported across 5 random seeds. 
    It demonstrates that our method outperforms baselines on three long-horizon surgical robot tasks. `-' indiates that the method completely failed on a certain task.
    } 
    \resizebox{1\linewidth}{!}{
 \begin{tabular}{ l c  c c c  c c c c c c  c c  c} 
 \toprule
  \multirow{2}{*}{\textbf{Method}} & & \multicolumn{3}{c}{\textbf{BiPegTransfer} with 3 subtasks and 100 episode steps} & & \multicolumn{3}{c}{\textbf{BiPegBoard} with 3 subtasks and 100 episodes steps} & & \multicolumn{3}{c}{\textbf{MatchBoardPanel} with 4 subtasks and 150 episode steps}  \\ [0.2ex]  \cmidrule{3-5}\cmidrule{7-9}\cmidrule{11-13} \cmidrule{11-13} 
    & & Succ. Rate ($\uparrow$) & Subtask Completion ($\uparrow$) & Rollout Len. ($\downarrow$) & & Succ. Rate ($\uparrow$) & Subtask Completion ($\uparrow$) & Rollout Len. ($\downarrow$) & & Succ. Rate ($\uparrow$) & Subtask Completion ($\uparrow$) & Rollout Len. ($\downarrow$) \\ [0.3ex] 
 \midrule [0.1ex]
  GCBC~\cite{ding2019goal} & & 11.27\ci{2.85} & 1.40\ci{0.05}& 76.26\ci{7.31} & & 8.26\ci{4.78} & 1.33\ci{0.08} & 80.22\ci{2.55} & & 5.50\ci{0.99} & 1.36\ci{0.03} & 124.47\ci{4.67} \\ [0.5ex]
 DEX~\cite{huang2023guided} & & 14.03\ci{1.45} & 1.55\ci{0.11} & 73.42\ci{8.24} & & 2.60\ci{1.15} &1.23\ci{0.04} & 81.76\ci{13.27} &  & - & -  & - \\ [0.5ex]
 T-STAR~\cite{lee2022adversarial} & & 67.73\ci{5.21} & 2.42\ci{0.09} & \textbf{65.04}\ci{8.58} & & 65.25\ci{9.95} & 2.37\ci{0.18} & \textbf{56.86}\ci{10.99} & & 45.42\ci{4.45} & 2.82\ci{0.09} & \textbf{94.77}\ci{3.64} \\ [0.2ex]
 \midrule
 \ourrow \algoName(Ours) & & \textbf{85.24}\ci{8.42} & \textbf{2.73}\ci{0.15} & \textbf{69.19}\ci{5.72} & & \textbf{81.76}\ci{4.08} & \textbf{2.67}\ci{0.07} & \textbf{61.21}\ci{5.22} & & \textbf{57.09}\ci{5.68} & \textbf{3.07}\ci{0.12} & \textbf{100.39}\ci{3.66} & \\ [0.4ex]
\bottomrule
\end{tabular}
}
\label{table:main_results}
\vspace{-0.4cm}
\end{table*}
\endgroup
\section{Experiments}
% In this section, we first conducted experiments on three long-horizon surgical robot tasks from the surgical simulation platform SurRoL~\cite{surrol}. Through extensive evaluations and model analysis, we aim to demonstrate the effectiveness of our proposed value-informed skill chaining to smoothly connect multiple skills for solving the entire task. Moreover, we also demonstrated the transferability of our learned policies to a dVRK platform in real-world tasks.  

\subsection{Environment Setup}
\subsubsection{Long-horizon surgical robot tasks} To demonstrate that our method is capable of solving long-horizon surgical robot tasks, we conducted experiments on the simulated platform SurRoL~\cite{surrol} with three tasks selected to be diverse and comprehensive, which cover a different number of surgical tools, various manipulation skills, and target objects, as illustrated in Fig.~\ref{fig:task}. Specifically, 1) the \textit{BiPegTransfer} task consists of three subtasks, including picking the block up from an initial peg with a patient-sided manipulator (PSM), handing the block over to another PSM, and finally placing the block at a target peg; the 2) the \textit{BiPegBoard} task shares a similar task structure with BiPegTransfer but additionally requires one PSM to orient the object in the last subtask; 3) the \textit{MatchBoardPanel} task is a single-manual surgical training task that requires one PSM to pull the panel door out, pick up the object, place it into a random cell on the board, and finally pull the panel door back.\footnote{In order to make this task more complex and cover more surgical skills, we designed a modified version with an additional sliding panel door.} All tasks are goal-conditioned whose state space is composed of object pose and robot proprioceptive state, and the action space is delta-position
control space, and the subgoal space is composed of the positions of the object and the end-effector. 
Following~\cite{lee2022adversarial}, 200 episodes of demonstrations collected by scripted controllers are provided to each subtask policy.

\subsubsection{Comparison methods} We compared our method against the following state-of-the-art methods: 1) \textbf{GCBC}~\cite{ding2019goal}, a representative imitation learning method that fits a parametric policy from demonstrations through a supervised objective and experience relabeling; 
2) \textbf{DEX}~\cite{huang2023guided}, an exploration-efficient demonstration-guided RL algorithm for surgical automation, which learns a single flat policy without task decomposition; 3) \textbf{T-STAR}~\cite{lee2022adversarial}, a representative skill chaining methods in robot manipulation tasks, which connects skills by regularizing the termination set to be covered by the next initiation set. While running its original implementation in the above tasks fails to make progress, which may be attributed to the sparse-reward setting, we customized this method in our framework by regularizing the terminal state to close to the initiation set of the next subtask policy.

\subsubsection{Implementation details} We adopted DEX for subtask policy learning. 
The learning rate of the actor and critic was set as 1e-4 and the coefficient of the exponential moving average was set as 5e-3 for stable training.
The hindsight experience replay was also adopted with a future sampling strategy. All subtask policies were trained until convergence, which takes around 2M environment steps. The same training steps are used to train baselines. In the skill chaining stage, we used SAC~\cite{sac} to learn both the value function and chaining policy. Specifically, we parameterized these two models as four-layer MLPs with ReLU activations, where each layer was of 256 hidden dimensions. The output of the actor was scaled to the task-specific range by a Tanh activation. All networks were trained with ADAM~\cite{adam} optimizer with a learning rate of 1e-4.

\subsection{Main Results}
We evaluated the manipulation performance with task success rate, the number of completed subtasks, and the total steps taken to accomplish the task. The evaluation is the average over five random seeds. We present the performance of our method and baselines on all three tasks in Table~\ref{table:main_results}. The results show that the imitation approach, GCBC, learns to mimic the skills in the early subtasks, such as picking up the object and pulling the panel door, but starts to deviate from the expected trajectory which often leads to the failed execution of skills in the later subtasks. This may be attributed to the poor generalization capabilities of such an imitation learning approach that requires a large number of demonstrations to overcome the distribution shift problem. 
Compared with imitation learning, the demonstration-guided RL approach, DEX, achieves a relatively higher success rate and execution efficiency in \textit{BiPegTransfer} task by using online experiences to address the distribution shift and demonstrations to overcome the exploration issue. However, it performs worse in the \textit{BiPegBoard} task, as the latter requires additional tool rotation before placing the ring to the target peg. Moreover, as the number of subtasks increases in the \textit{MatchBoardPanel}, the DEX agent is more struggling with collecting successful trials and fails to make progress.

\begin{figure}[t]
    \centering
    \vspace{0.2cm}
    \centerline{\includegraphics[width=1\linewidth]{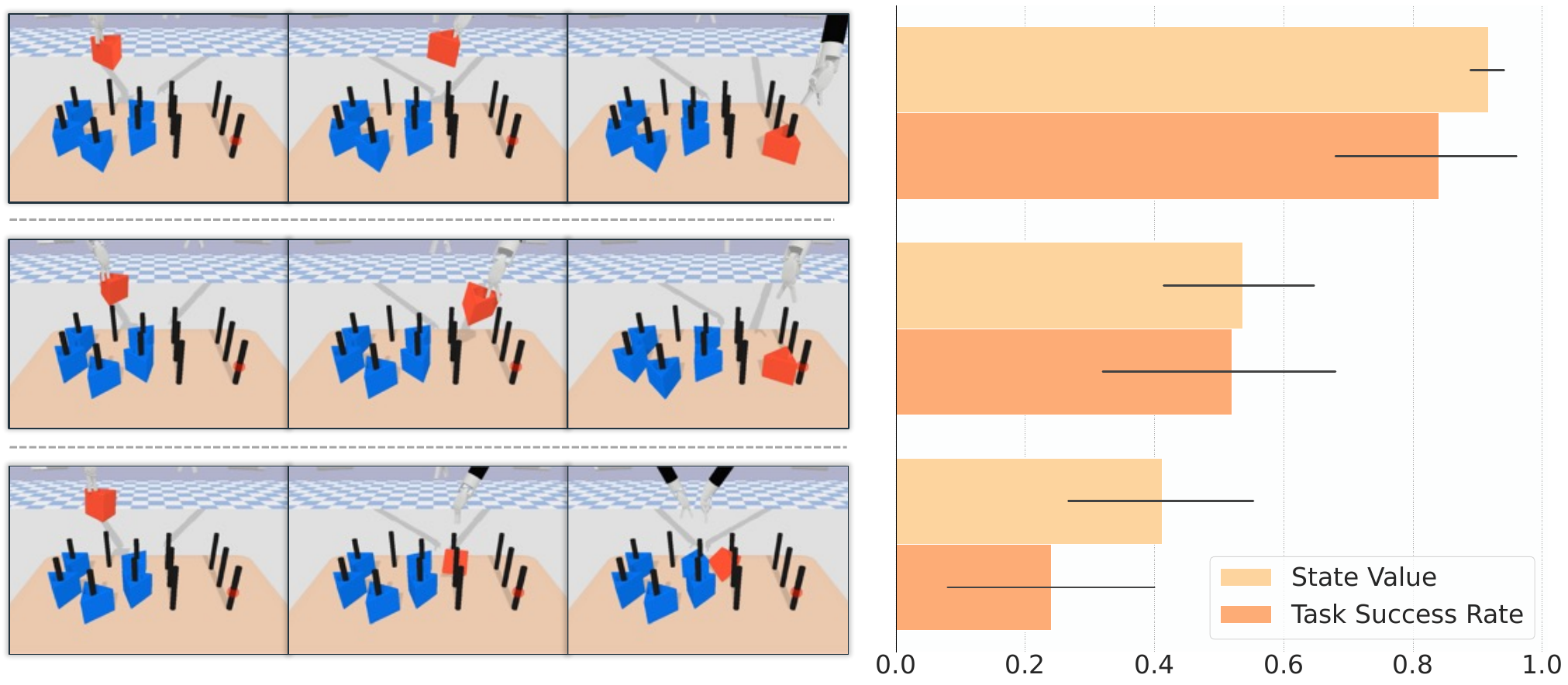}}\vspace{-1mm}
    \caption{\textbf{State values and corresponding task success rates.} We collect three sets of trajectories (each of 25 episodes) with different levels of values that estimate the expected success probability of the whole task.
    The results show that our value-informed approach is able to distinguish states in the terminal set of the first subtask and thus instruct subtask policies to smoothly connect with each other for accomplishing the whole task.}
    \label{fig:visualization}
    \vspace{-0.4cm}
\end{figure}

\begin{figure*}[t!]
    \centering
    \centerline{\includegraphics[width=1\linewidth]{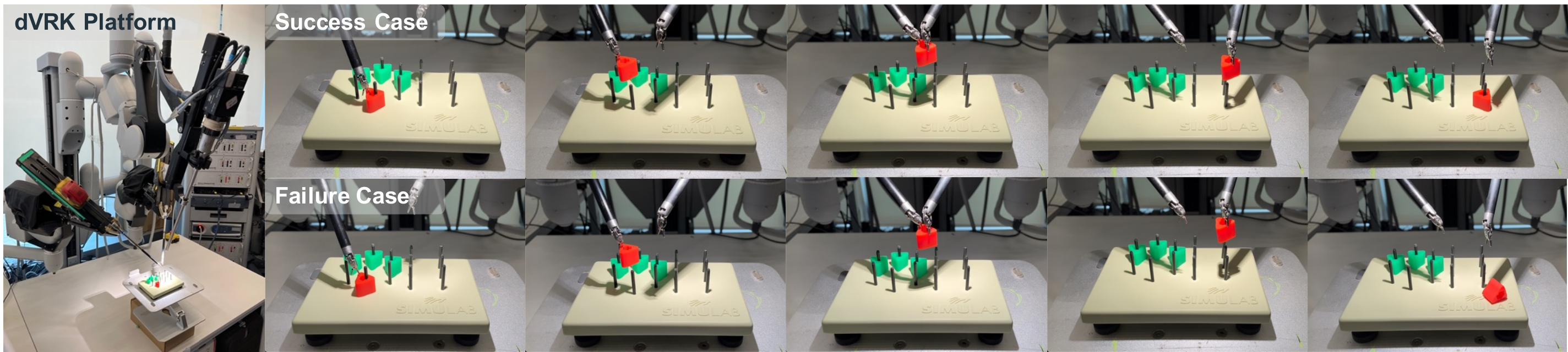}}
    \caption{\textbf{Real-world deployment on the dVRK platform.}
    We learn the policy in SurRoL and deploy the trajectories generated by the learned policies to the real-world bimanual peg transfer task with a surgical robot of the dVRK platform. We test 20 trials of task execution in total and get 16 successful trajectories, which demonstrates the transferability of our method in real-world surgical robot tasks with long-horizon structures.}
    \label{fig:dvrk}
    \vspace{-0.6cm}
\end{figure*}

On the other hand, the prior skill chaining approach, T-STAR, explicitly harnesses the compositional structure of the long-horizon tasks by learning multiple skills to handle each subtask, achieving higher success rates in three tasks than in previous baselines. However, it often fails to chain the skills in the later stage of the entire task as the connection between current skills does not ensure the future connections will be successfully performed, which is represented by subtask completion in Table~\ref{table:main_results}. Instead, our method ensures more smooth connections of all subsequent skills when doing the current chaining, which achieves significant performance improvement on all tasks. Meanwhile, the results also show that, compared with T-STAR, our method takes slightly more steps to accomplish the task. We observe that our methods will make some tiny actions to mildly adjust the terminal state even when the subtask has been successfully executed. This verifies the effectiveness of our value-informed chaining approach that terminating each skill at states with higher values makes the overall subtask connections more smooth.

\subsection{Ablation Study on Skill Chaining Strategy}
Essentially, our value-informed skill chaining approach takes a long-term view when connecting two skills by evaluating the expected success probability of the entire task, while prior methods take a relatively short-term strategy by matching the current initiation set and termination set. To this end, we aim to verify the effectiveness of our choice of value functions. We introduce three variants of our methods that take different value functions: 1) {\algoName-DM} that evaluates a terminal state of one subtask policy with a discriminator discerning whether the terminal states fall inside the termination set; 2) {\algoName-LDM} that extends the value function in \algoName-DM to measure whether all subsequent terminal states will fall insides their target terminal sets; 3) {\algoName-SR} that evaluates the terminal with subtask reward instead of the reward of the entire task. The results in Table~\ref{table:abl} show that, compared with \algoName-DM, \algoName-LDM achieves $9.97\%$ and $13.37\%$ performance improvement on the \textit{BiPegTransfer} and \textit{BiPegBoard}, respectively. This not only demonstrates that a value function with a long-term view makes the policy connections more smooth, but also our method is of good extensibility to existing skill chaining methods. Such a conclusion can be further verified by the observation that, compared with \algoName, \algoName-SR also exhibits a performance drop on both tasks due to its short-term evaluation. Surprisingly, we observe that \algoName-SR also underperforms \algoName-DM, indicating that evaluating state with subtask success probability is more likely to incur a terminal state that falls outside the next initiation set, while the distribution matching can ensure a successful connection between the current two subtask policies.

% \begingroup
% \setlength{\tabcolsep}{4pt}
% \begin{table}[t]
% \vspace{7mm}
% \caption{\textbf{Ablations.} ...}
% \label{table:ablations}
% \centering
% \resizebox{0.5\textwidth}{!}{%
% \begin{tabular}{lccccccccc}
% \toprule
% \multirow{2}{*}{\textbf{Method}} & & \multicolumn{3}{c}{\textbf{BiPegTransfer} & & \multicolumn{3}{c}{\textbf{BiPegBoard} \\ 
% % \textbf{Method} & \textbf{Metrics} & T-STAR &  VISC-SR & VISC-DM & VISC  \\\midrule
% % BiPegTransfer & Succ. Rate  & 67.73\ci{5.21} &  & 76.21\ci{0.07} & 85.24\ci{8.24}  \\%& 0.45\\
% & Avg. Value &\\ [0.3ex]
% \cdashline{1-6}\noalign{\vskip 0.6ex}
% BiPegBoard & Succ. Rate & 65.25\ci{9.95} &  & 77.50\ci{0.08} & 81.76\ci{4.08} \\%& 0.75\\
% & Avg. Value \\
% % \midrule
% % \ourmethod~(ours)   & $\mathbf{0.90}$ & $\mathbf{0.95}$  & $\mathbf{0.75}$ & $\mathbf{0.90}$  & $\mathbf{1.00}$ \\%& $\mathbf{0.90}$\\
% \bottomrule
% \end{tabular}}

\begingroup
\setlength{\tabcolsep}{3.5pt}
\begin{table}[t]
    \centering
    \vspace{0.2cm}
    \caption{\textbf{Results of ablation study.} We test three variants of \algoName to investigate the effect of different choices of the value function. The results show that our proposed one achieves the best performance by using the environment reward of the entire task.
    } 
    \resizebox{1\linewidth}{!}{
 \begin{tabular}{ l c c c c c} 
 \toprule
 \multirow{2}{*}{\textbf{Variant}} & \multicolumn{2}{c}{\textbf{BiPegTransfer}} & & \multicolumn{2}{c}{\textbf{BiPegBoard}} \\ [0.2ex] \cmidrule{2-3}\cmidrule{5-6} & Succ. Rate ($\uparrow$) & Subtask Completion ($\uparrow$) & & Succ. Rate ($\uparrow$) & Subtask Completion ($\uparrow$) \\
 \midrule [0.2ex]
    \algoName-DM & 66.23\ci{5.20} & 2.37\ci{0.09} & & 64.13\ci{9.95} & 2.23\ci{0.15}\\ [0.5ex]
    \algoName-LDM & 76.20\ci{6.56} & 2.56\ci{0.10} & & 77.50\ci{7.90} & 2.59\ci{0.16}\\ [0.5ex]
    \algoName-SR & 61.42\ci{4.36} & 2.22\ci{0.06} & & 60.25\ci{6.29} & 2.10\ci{0.07}\\ [0.1ex]
    \midrule 
    \ourrow \algoName (Ours) & \textbf{85.24}\ci{8.42} & \textbf{2.73}\ci{0.15} & & \textbf{81.76}\ci{4.08} & \textbf{2.67}\ci{0.07} \\
\bottomrule
\end{tabular}
}
\label{table:abl}
\vspace{-0.3cm}
\end{table}
\endgroup

\subsection{Analysis of Learned State Value Function}
We validate that our value function can accurately estimate the expected success probability of the whole task and accordingly instruct subtask policies to connect with each other smoothly. Specifically, three sets of different levels of state value are collected. The results in Fig.~\ref{fig:visualization} show that the empirical task success rate of each trajectory set is close to the estimated value. For instance, the trajectory with a low value of terminal state (left bottom) completes the first subtask, but fails to complete the following subtask tasks due to the insufficient contact between the tip of a surgical tool and the object at the end of the first subtask, which makes the handing over difficult. While the trajectory with intermediate value (left middle) smoothly connects the first two subtasks, the orientation of the object after the handing over leads to the failure of the final subtask that requires proper orientations to place the object to the target peg. In contrast, the trajectory generated by our learned policy is of a high value (left upper) and successfully connects all subtask policies, indicating the effectiveness of our accurate value estimates for skill chaining.

\subsection{Deployment on the dVRK Platform}
We conducted experiments on the real dVRK platform to validate the transferability of our trained policies. Specifically, the representative bi-manual task \textit{BiPegTransfer} is selected to be automated, which covers a wide range of complex skills. We run the best-performing policies of our method on the simulated environments and collect the generated trajectories for deployment, without extra real-robot learning. Each PSM is associated with 4 degrees of freedom (DoF), which consists of 3-DoF of translation and 1-DoF of rotation, and one action for controlling the jaw. The snapshots of the automation are shown in Fig.~\ref{fig:dvrk}. During the deployment, we find that the robot is able to sequentially perform the acquired skills of both single-hand manipulation and bimanual coordination as we observed in the simulator. We totally test 20 trials of task execution and observe 16 successful trials. This demonstrates the potential of our method in solving long-horizon surgical robot tasks on the real robot.

\section{Conclusion and Future Work}
In this work, we present ViSkill, a novel value-informed skill chaining framework for policy learning of long-horizon surgical robot tasks. ViSkill learns subgoal-based subtask policies in decomposed long-horizon tasks, and subsequently learns a chaining policy associated with a value function to smoothly connect these policies for accomplishing the entire task.
We demonstrate the effectiveness of our method on three long-horizon surgical robot tasks from the surgical simulation platform, where significant improvements in task success rates have been achieved. We also demonstrate the deployment of the learned policies on the dVRK hardware.

During our real-world deployment, we observed a few instances of failure, which we attribute to the exacerbation of the sim-real gap in long-horizon tasks. For instance, in Fig.~\ref{fig:dvrk}, a slight orientation of the block led to failed task execution. This motivates further investigation of sim-to-real adaptation methods in future work for advancing the automation of surgical robots. Furthermore, investigating the possibility of reducing the hand-crafted design to decompose tasks is also desirable to further improve the scalability of the system.
% \addtolength{\textheight}{-12cm}   % This command serves to balance the column lengths
%                                   % on the last page of the document manually. It shortens
%                                   % the textheight of the last page by a suitable amount.
%                                   % This command does not take effect until the next page
%                                   % so it should come on the page before the last. Make
%                                   % sure that you do not shorten 

% \clearpage
% \clearpage
{

\footnotesize
\bibliographystyle{IEEEtranN}
\bibliography{ref}

}

\end{document}